\documentclass{article}
\usepackage{spconf,amsmath,epsfig}
\usepackage[noend]{algpseudocode}
\usepackage{algorithmicx,algorithm}

\usepackage{fancyhdr}
\begin{document}\sloppy

\def\x{{\mathbf x}}
\def\L{{\cal L}}

\title{DYNAMIC REGION DIVISION FOR ADAPTIVE LEARNING PEDESTRIAN COUNTING}

%
\name{Gaoqi He$ ^1$, Zhenwei Ma$ ^2$, Binhao Huang$ ^2$, Bin Sheng$ ^3$, Yubo Yuan$ ^2$}
\address{$^1 $East China Normal University, $^2 $East China University of Science and Technology, \\
$^3 $Shanghai Jiao Tong University\\
hegaoqi@sei.ecnu.edu.cn, 960901625@qq.com, 805493736@qq.com, \\shengbin@cs.sjtu.edu.cn, ybyuan@ecust.edu.cn
\thanks{This work was supported by National Key Research and Development Program (No.2016YFA0502304), the Open Project Program of State Key Laboratory of Virtual Reality Technology and Systems, Beihang University (No.VRLAB2018B12), the Open Project Program of the State Key Lab of CAD\&CG, Zhejiang University (No.A1913).}
}

\maketitle

\begin{abstract}
Accurate pedestrian counting algorithm is critical to eliminate insecurity in the congested public scenes. However, counting pedestrians in crowded scenes often suffer from severe perspective distortion. In this paper, basing on the straight-line double region pedestrian counting method, we propose a dynamic region division algorithm to keep the completeness of counting objects. Utilizing the object bounding boxes obtained by YoloV3 and expectation division line of the scene, the boundary for nearby region and distant one is generated under the premise of retaining whole head. Ulteriorly, appropriate learning models are applied to count pedestrians in each obtained region. In the distant region, a novel inception dilated convolutional neural network is proposed to solve the problem of choosing dilation rate. In the nearby region, YoloV3 is used for detecting the pedestrian in multi-scale. Accordingly, the total number of pedestrians in each frame is obtained by fusing the result in nearby and distant regions. A typical subway pedestrian video dataset is chosen to conduct experiment in this paper. The result demonstrate that proposed algorithm is superior to existing machine learning based methods in general performance.
\end{abstract}


\section{Introduction}
\label{sec:intro}

Crowd gathering lead to huge loss of people's life, property and cause terrible social influence~\cite{liu2016research}. Subway station, railway station and scenic area are typical crowd gathering scene. In Shanghai, China, the daily average passenger volume in subway station comes to 10,330,000 in 2018 and increases up more than 11\% from a year earlier. Therefore, the research of methods to avoid crowd gathering has important application value for public management.

\begin{figure}[t]
\begin{minipage}[b]{1.0\linewidth}
  \centering
  \centerline{\epsfig{figure=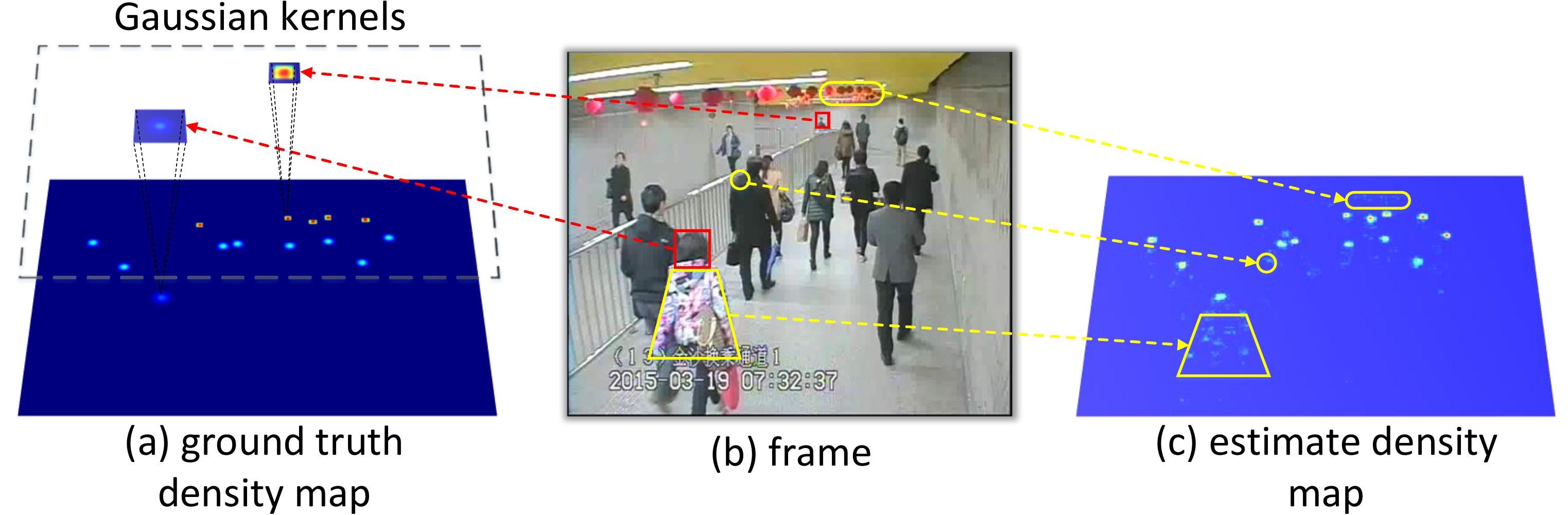,width=8.5cm}}
\end{minipage}
\caption{Two issues about density-based methods in subway scenes. Red line: the error of gaussian kernel simulation. Yellow line: misidentification of clutter background. Best viewed in color.}
\label{fig:fig1}
\end{figure}

\begin{figure*}[t]
\begin{minipage}[b]{1.0\linewidth}
  \centerline{\epsfig{figure=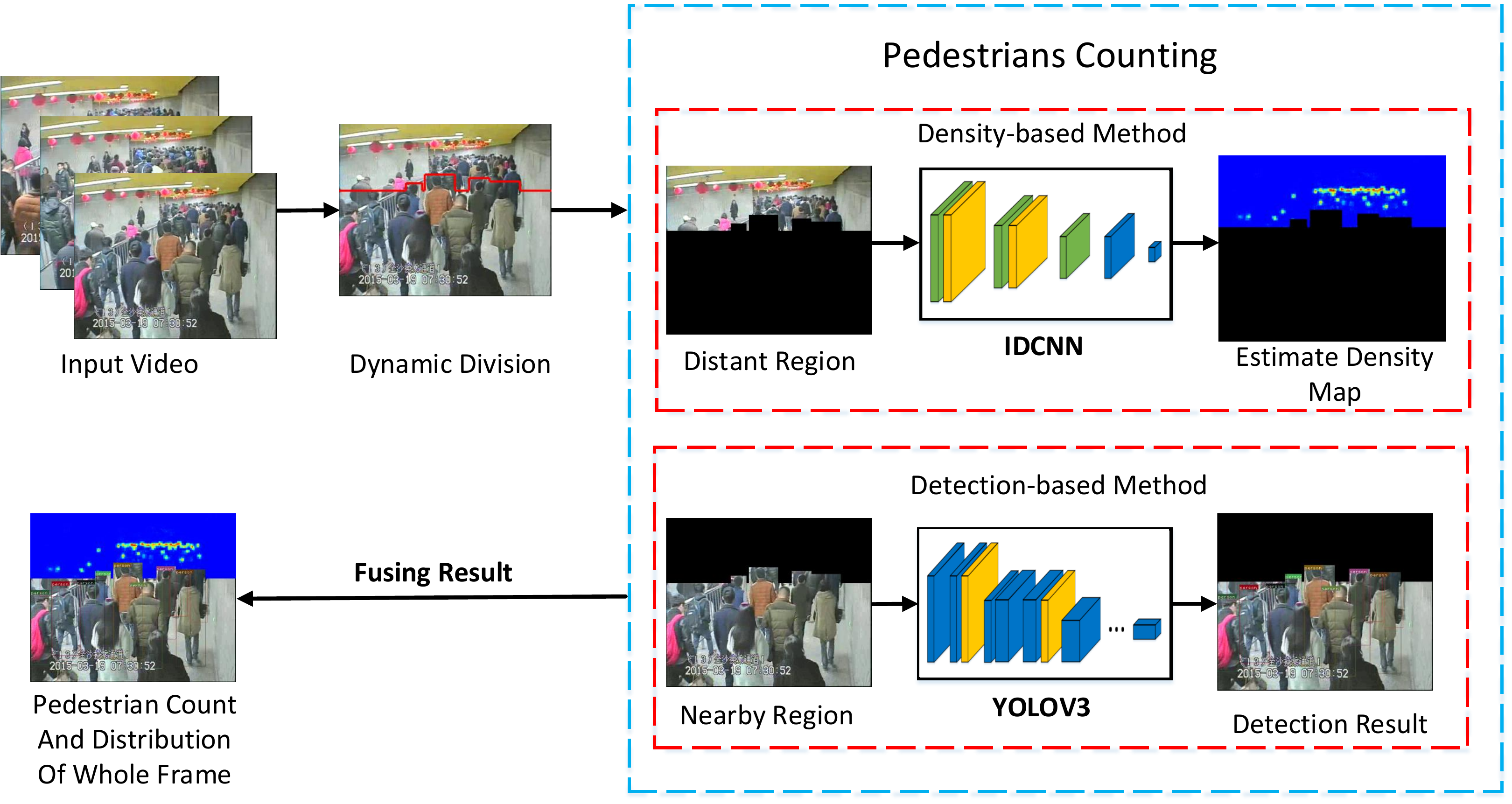,width=14cm}}
\end{minipage}
\caption{Overview of Dynamic Region Learning algorithm.}
\label{fig:fig2}
\end{figure*}

Pedestrian counting plays an essential role in crowd monitoring~\cite{lempitsky2010}. In the past years, computer vision and machine learning methods have been widely applied to the field of pedestrian counting. Researchers have proposed various pedestrian counting methods which can be divided into three categories~\cite{zhao2017learning}: detection-based methods~\cite{dalal2005histograms}~\cite{redmon2016you}, regression-based methods~\cite{fiaschi2012}~\cite{he2017double} and density-based methods~\cite{zhang2015}~\cite{zhang2016}. Detection-based methods locate objects accurately but often suffer from scale variance, clutter backgrounds and occlusions. Regression can get more accurate results but hard to design a good feature representation. Recently, density-based methods use a two-dimension gaussian kernel to simulate pedestrian head and have demonstrated strong performance in extremely crowded and large view angle scenes such as ShanghaiTech~\cite{zhang2016} and WorldExpo~\cite{zhang2015}. As shown in Figure~\ref{fig:fig1}, density-based methods have two main shortcuts in high perspective distortion scenes. One is the error of gaussian kernel simulation and the other is misidentification of clutter background caused by perspective distortion. In Figure~\ref{fig:fig1}, red lines reflect the error of gaussian kernel simulation. Due to perspective distortion, gaussian kernels with different sizes and sigma setting are established to simulate the heads in different regions of scene. It is effective for the heads far away from camera. While the head is close to camera, it is not appropriate due to complicated texture such as women's long hair, collars, and hats. As a result, neural network can not learn a good counting result for the region close to camera. Yellow lines in the Figure~\ref{fig:fig1} show that density-based methods misidentify the clutter background as head such as clothes' texture, bags' corner and pedestrians' elbow angle. The reason to this phenomenon is that heads with multiple scales are trained together and neural networks learn an average representation for heads of all sizes. Therefore, using density-based method in high perspective distortion scenes can not obtain a optimal counting result. All the aforementioned methods consider only one method for entire frame, while He \emph{et al.}~\cite{he2017double} proposed a straight-line double region pedestrian counting method. It make the best use of features from different regions to design appropriate counting method. However, using a straight line may cut one head into two parts which lead to the error counting. Driven by this work, we propose a dynamic region division algorithm to keep the completeness of counting objects. Contributions of this paper are summarized as follows: 1) Utilizing the object bounding boxes obtained by YoloV3 and expectation division line of the scene, the boundary for nearby region and distant one is generated under the premise of keeping the completeness of heads. 2) Appropriate learning models are applied to count pedestrians in each obtained region. In distant region, a novel inception dilated convolutional neural network is proposed to solve the problem of choosing dilation rate. In nearby region, YoloV3 is used to detect pedestrian in multi-scale.

\section{RELATED WORK}

\textbf{Counting by density map} Density map was first introduced into pedestrian counting field by Lempitsky et al.~\cite{lempitsky2010} and used 2D gaussian kernel to model one pedestrian. Then Fiaschi et al.~\cite{fiaschi2012} used random forest to regress the object density and improved training efficiency. With the powerful ability of deep learning, Zhang \emph{et al.}~\cite{zhang2015} first explored deep models for crowd counting and used two 2D gaussian kernels to model pedestrian's head and body separately. Huang \emph{et al.}~\cite{huang2018} considered that it is inaccurate to use a gaussian kernel to model pedestrian's body. They applied semantic segmentation to extract body part instead and achieved more accurate counting result. However, other methods abandoned the modeling of body and only modeled pedestrian head. Zhang \emph{et al.}~\cite{zhang2016} proposed a geometry-adaptive method to generate proper kernel for different head sizes, but it was only suitable for extremely crowd scenes. Besides, multi-column CNN(MCNN) was introduced to use filters of different sizes to model the density maps corresponding to heads of different scales. Following this work, later methods mainly focused on the improvement of network structure and added more extra information to network. Li \emph{et al.}~\cite{li2018csrnet} conducted an experiment to show that multi-column structure was inferior to a deeper network. Information of people number was widely added to network through various schemes~\cite{sam2018divide}~\cite{sindagi2017cnn}. Zhao \emph{et al.}~\cite{zhao2017learning} embedded perspective information into deconvolution network. These information raised the counting accuracy but still used gaussian kernels to model heads with larger size.

\textbf{Counting by detection} Traditional methods used the histogram of oriented gradients(HOG) as the pedestrian-level features and the support vector machine as the classifier to detect pedestrians in specific scenes~\cite{dalal2005histograms}, but these hand-crafted features severely suffered from light variance and scale variance. The region-based convolutional neural networks(R-CNNs)~\cite{girshick2014rich} used features extracted from CNN and improved the performance in detection. This method could be summarized as two stages processing: proposal and classification, but hard to be accelerated. YOLO~\cite{redmon2016you} provided a new one-stage solution for detection and significantly improved the speed. It converted the thought of classification to regression in sub-grids and abandoned the process of proposal. Following YOLO, some methods paid attention to support multiple scales object detection such as SSD~\cite{liu2016ssd} and YOLOV3~\cite{redmon2018yolov3}. Although detection methods have achieved tremendous performance and can be used in sparse crowd scenes, it is hard to substitute density-based methods in crowded scenes.

\section{Dynamic region division algorithm}

\subsection{Overview}
To overcome the error of gaussian kernel simulation and misidentification of clutter background caused by perspective distortion, a novel algorithm framwork is proposed. Figure~\ref{fig:fig2} clearly shows the flow chart of algorithm. Since we find that gaussian kernels are not suitable for simulating large heads, basing on the straight-line double region pedestrian counting method~\cite{he2017double}, we propose a dynamic region division algorithm to keep the completeness of counting objects. Utilizing the object bounding boxes obtained by YoloV3 and expectation division line of the scene, the boundary for nearby region and distant one is generated under the premise of retaining whole head. Then in the nearby region, we apply YoloV3 detector to detect pedestrians that can avoid occurrence of identifying clutter background as heads. In the distant region, we introduce dilated convolution layer into our density map based network to enlarge the receptive field and further design an inception module to address the problem how to choose dilation rate. Finally, we fuse the counting results from two parts and also obtain the total distribution information.

\begin{algorithm}[t]
\caption{dynamic mask generation} 
\hspace*{0.02in} {\bf Input:} 
expectation height \emph{H}, pedestrian bounding boxes $b_{i}$\\
\hspace*{0.02in} {\bf Output:} 
division mask \emph{M}
\begin{algorithmic}[1]
\State/*\emph{M} is initialized as a zero matrix with the same rows and columns as original frame*/ 
\State/*\emph{n} denotes the total number of bounding boxes*/
\For{$i$ = 1 to \emph{n}} 
　　\If{$i$ == 1} 
        \State /*fill mask \emph{M} with 1 in items from \emph{H} to $I_{height}$ rows and \emph{0} to $tf_{i}^{x}$ columns*/
　　　　\State $M$[$H$:$I_{height}$, $0$:$tf_{i}^{x}$] = 1
　　\Else
        \State /*current bounding box overlap last box*/
        \If{$tf_{i}^{x}$ $<=$ $br_{i-1}^{x}$}
            \State $M$[$tf_{i-1}^{y}$:$I_{height}$, $tf_{i-1}^{x}$:$tf_{i}^{x}$] = 1
        \Else
            \State $M$[$tf_{i-1}^{y}$:$I_{height}$, $tf_{i-1}^{x}$:$br_{i-1}^{x}$] = 1
            \State $M$[$H$:$I_{height}$, $br_{i-1}^{x}$:$tf_{i}^{x}$] = 1
        \EndIf
　　\EndIf
\EndFor
\State /*fill the mask in the right side of last bounding box*/
\State $M$[$tf_{i-1}^{y}$:$I_{height}$, $tf_{i-1}^{x}$:$br_{i-1}^{x}$] = 1
\State $M$[$H$:$I_{height}$, $br_{i-1}^{x}$:$I_{width}$] = 1
\State \Return mask \emph{M}
\end{algorithmic}
\end{algorithm}

\begin{figure}[t]
\begin{minipage}[b]{1.0\linewidth}
  \centerline{\epsfig{figure=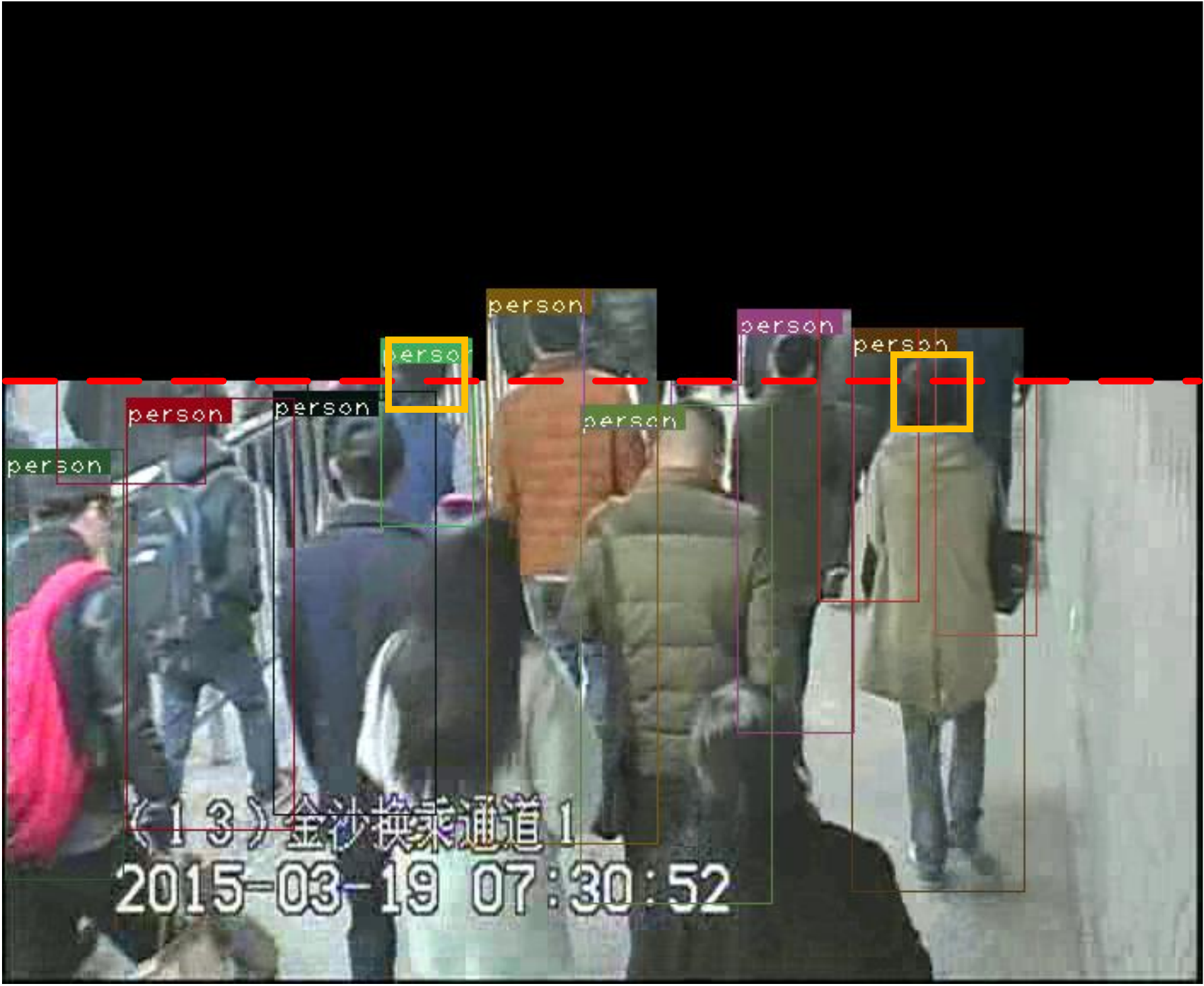,width=3.5cm}}
\end{minipage}
\caption{Visualization of Dynamic Region Division.}
\label{fig:fig3}
\end{figure}

\subsection{Dynamic region division}
To maximize advantages of different methods in corresponding regions, it is significant to divide regions properly. He \emph{et al.}~\cite{he2017double} used a straight line to divide, but it causes the error that one head may be cut into two parts by the line. To avoid this problem, we propose a dynamic region division algorithm. The detailed steps are described below.

(1) For each frame $I_{k}$ in surveillance video \emph{V}= \{$I_{1}$,$I_{2}$,...,$I_{M}$\} with a resolution of $I_{width}$ * $I_{height}$, YoloV3 detector is used to detect the pedestrians. We record the position of each detected pedestrian's center $s_{i}$ in all frames of the video as
\begin{eqnarray}
s_{i} &=& (s^{x}_{i},s^{y}_{i})
\end{eqnarray}

(2) In order to calculate the distribution of the detected pedestrian along the frame height, we count the number of detected pedestrian in each height $h_{i}$ (from bottom to top) and record the number as $k_{i}$. Then we obtain the possibility $p_{i}$, which denotes the possibility of the pedestrian detection in height $h_{i}$
\begin{eqnarray}
p_{i} &=& k_{i}/\sum_{j=1}^{I_{height}}k_{j}
\end{eqnarray}

(3) Then we calculate the expectation \emph{H} as the height for irregular region division later.
\begin{eqnarray}
H &=& \sum_{i=1}^{I_{height}}p_{i}h_{i}
\end{eqnarray}

(4) To avoid cutting one head into two parts, we calculate a dynamic division mask based on detected pedestrians and expectation height \emph{H}. For each detected pedestrian bounding box $b_{i}=(tf_{i},br_{i})$, we record its top left corner $tf_{i}=(tf_{i}^{x},tf_{i}^{y})$ and bottom right corner $br_{i}=(br_{i}^{x},br_{i}^{y})$. We first find pedestrian bounding boxes whose head part height $(1-\alpha)tf_{i}^{y} + \alpha br_{i}^{y} < H$ means that these heads will be divided into two parts if use a straight line. $\alpha$ is the proportion of head to the whole body and we set 0.3 in our experiment. Then we input these boxes into algorithm 1 to get a mask, where 1 in mask represents distant region and 0 denotes nearby region. We can obtain the regions through mask. Figure~\ref{fig:fig3} shows an example of dynamic region division. Heads with yellow rectangle denote that they are cut into two parts by using straight red line~\cite{he2017double}, but our dynamic region division can effectively keep the completeness of heads.

\section{Counting model}

\subsection{Counting model for distant region}
\begin{figure}[t]
\begin{minipage}[b]{1.0\linewidth}
  \centering
  \centerline{\epsfig{figure=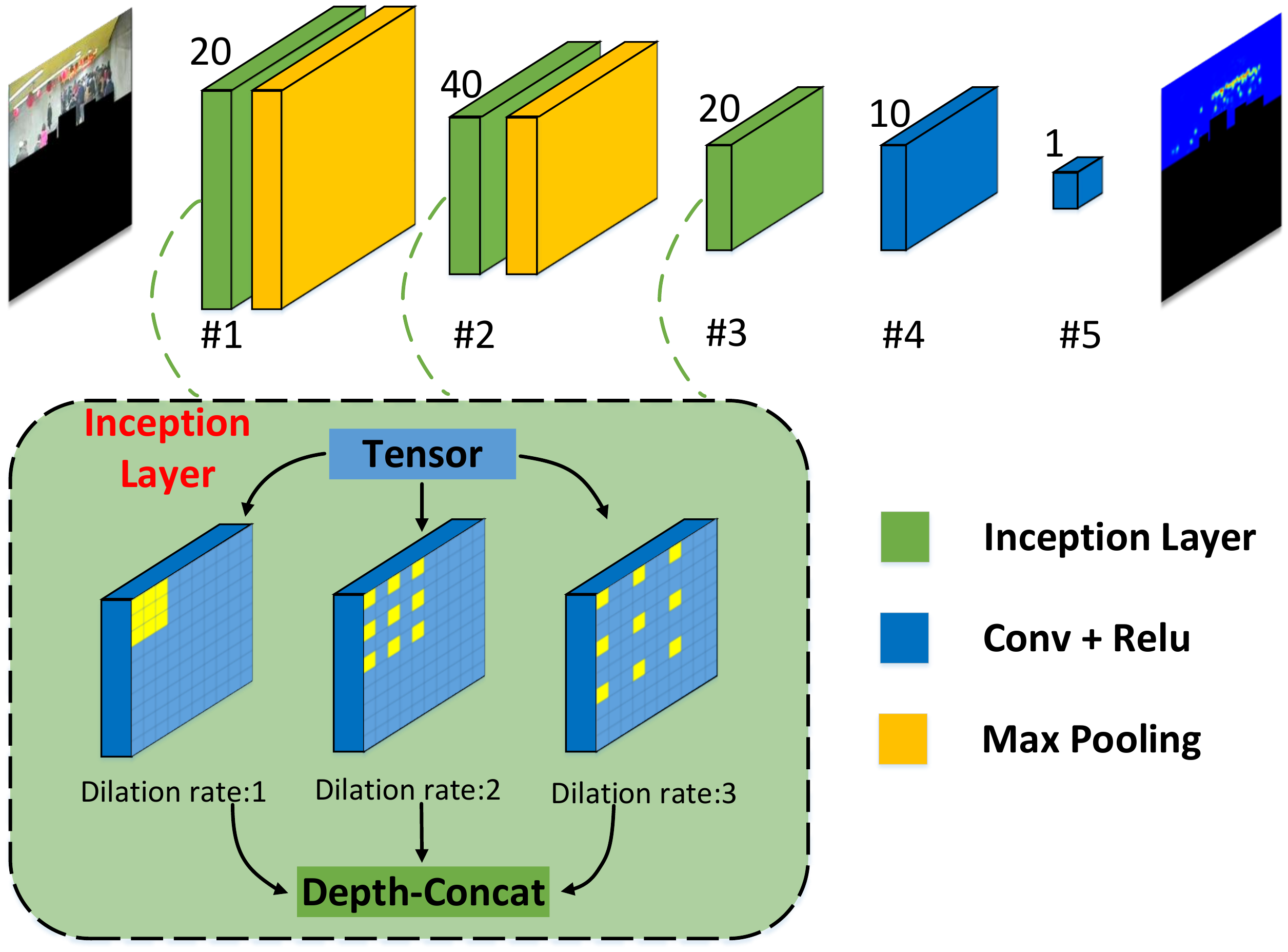,width=7.5cm}}
\end{minipage}
\caption{The structure of IDCNN.}
\label{fig:fig4}
\end{figure}

Li \emph{et al.}~\cite{li2018csrnet} proposed a CSRNet for crowd counting which introduced dilated convolution to improve traditional convolutional neural network. Dilated convolution enlarges the receptive field without increasing the number of parameters or the amount of computation. However, it is hard to choose the dilation rate and CSRNet prepared four dilation rate configurations to decide final dilation rate according to the performance. Inspired by Szegedy \emph{et al.}~\cite{szegedy2015going}, we propose an inception layer to address the problem of dilation rate chosen as shown in Figure~\ref{fig:fig4}. The main idea is that instead of needing to pick one of these dilation rates, we can concatenate all the outputs and let the network learn whatever parameters it wants to use. There are three dilated convolution kernels with same kernel size but different dilation rate in our inception layer. Following ~\cite{li2018csrnet}, we choose 1,2,3 as the dilation rate for three kernels. We then concatenate these three outputs in depth channel. The upper part of Figure~\ref{fig:fig4} shows our IDCNN in detail. There are three inception layer in our network and a max pooling layer behind first two inception dilated module. We doesn't remove max pooling because it is harder for a fully convolutional neural network to converge with original resolution than downsampled resolution. In our network, all the convolution kernel size is 3$\times$3. After the third inception dilated module layer, we use a convolution layer with 2 dilation rate and 1$\times$1 convolution to generate the density map.

In the distant region, head scale is small and it is suitable for using gaussian kernel to simulate head. For an input RGB frame, we aim to output a density map~\cite{lempitsky2010}. The ground truth density map is created as:
\begin{eqnarray}
D_{i}(p) &=& \sum_{P\in \textbf{P}_{i}} \frac{1}{\|\textbf{P}_{i}\|} \mathcal N (p;P,\sigma)
\end{eqnarray}

where $D_{i}(p)$ denotes the density value of pixel $p$ in density map corresponding to $i$-th frame. $P$ is the center position of pedestrian head while $\textbf{P}_{i}$ is the collection of all annotated head centers. A normalized 2D Gaussian kernel $\mathcal N$ is used to model a head with variance $\sigma$. $\|\textbf{P}_{i}\|$ is the number of annotated heads and the whole distribution is normalized by its reciprocal. Besides, to better simulate the head, $\sigma$ is related to perspective map $M(p)$. Manual annotation was used to get $M(p)$ in ~\cite{zhang2015}, but we take the advantage of detector in the nearby region and use linear regression to calculate $M(p)$ with all the detected bounding boxes. We set $\sigma = 0.15M(p)$.

\begin{table*}[t]
\begin{center}
\caption{Mean absolute errors of the subway station pedestrian dataset} \label{tab:cap}
\begin{tabular}{|c|c|c|c|c|c|c|}
  \hline
  Methods & Jinshajiang road & Jing'an temple & South railway station & People square & Xujiahui & Average
  \\
  \hline
  He \emph{et al.}~\cite{he2017double} & 2.96 & 1.66 & 2.82 & 3.90 & 2.42 & 2.75 \\
  \hline
  MCNN~\cite{zhang2016} & 2.17 & 1.44 & \textbf{2.35} & 5.35 & 2.13 & 2.69 \\
  \hline
  CSRNet~\cite{li2018csrnet} & 2.10 & \textbf{1.43} & 2.76 & 5.65 & 2.16 & 2.82 \\
  \hline
  Ours & \textbf{2.07} & 1.59 & 2.81 & \textbf{3.41} & \textbf{1.61} & \textbf{2.30} \\
  \hline
\end{tabular}
\end{center}
\end{table*}

\begin{figure}[t]
\begin{minipage}[b]{1.0\linewidth}
  \centering
  \centerline{\epsfig{figure=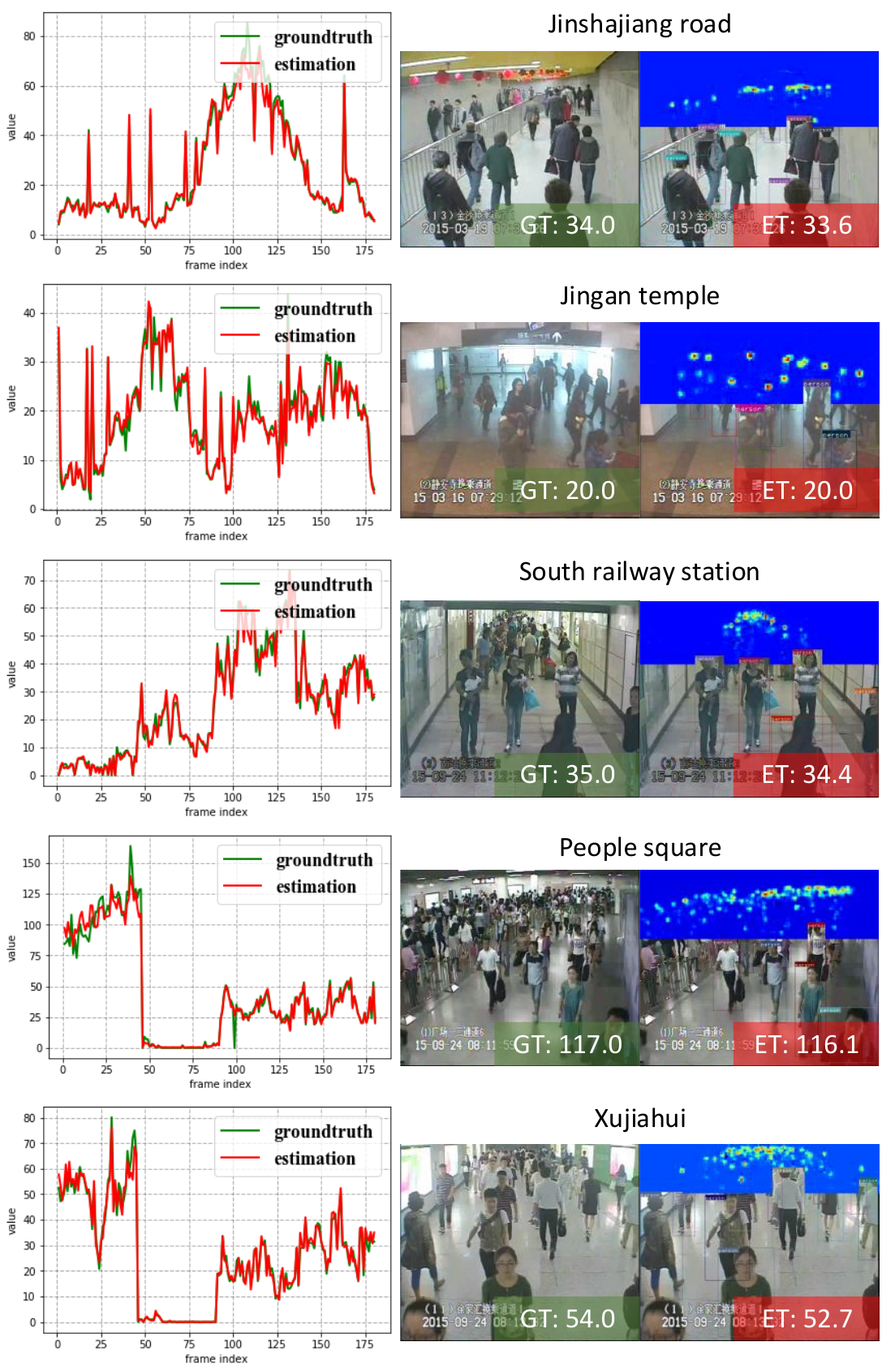,width=7.5cm}}
\end{minipage}
\caption{Counting and distribution estimation results on the subway station pedestrian dataset.}
\label{fig:fig5}
\end{figure}

\subsection{Counting model for nearby region}
In the nearby region, pedestrian has detailed pedestrian-level features. Therefore, it is not accurate to use gaussian kernel to simulate head and we use the detection-based method to count pedestrians instead. Recently Redmon \emph{et al.}~\cite{redmon2018yolov3} have proposed YOLOV3 detection method. It is an incremental improvement of YOLO and support multi-scale detection which is suitable for pedestrian detection in our nearby regions. Since we only need to detect pedestrian, we modify the output layer and only reserve three anchor boxes with following structure in each grid: [$p_{o},t_{x},t_{y},t_{w},t_{h}$]. $p_{o}$ is the probability of detected pedestrian. $t_{x},t_{y}$ is the center of pedestrian and $t_{w},t_{h}$ is the width and height of the bounding box. People-labeled data in COCO dataset~\cite{lin2014microsoft} are used to train this modified detector.

\section{Experiment}

\subsection{Experiment dataset}
We evaluate the proposed algorithm through extensive experiments on the publicly available Subway station pedestrian dataset~\cite{he2017double}. The dataset covers five typical subway station scenes in Shanghai. These scenes mainly locates at the transfer corridors, which have severe perspective distortion and large variance of head scale. Since we use density-based method in the distant region, it is necessary to annotate each head position in frames. Therefore, we add the annotation of head positions to original dataset. For each scene, there are 420 frames for training and 180 frames for testing. The average pedestrian count is 28.78.

\subsection{Model training}
\textbf{Training} Due to our division algorithm, distant region is dynamic. Therefore, it can not be the input of IDCNN directly and we fill zero to make region be a rectangle instead. Since there are two max pooling layers in IDCNN, ground-truth density map was downsampled to 1/4 of the original height and width. To augment the training set for training the IDCNN, we flip the frames horizontally to double the training set. Following ~\cite{he2017double}, we use the absolute error($MAE$) as the evaluation metric which are defined as follows:
\begin{eqnarray}
MAE &=& \frac{1}{N} \sum_{i=1}{N} \left|z_{i}-z_{i}^{gt}\right|
\end{eqnarray}
where $N$ is the number of test images, $z_{i}$ is the estimated number of pedestrians in the $i$-th frame, and $z_{i}^{gt}$ is the actual number of pedestrians in the $i$-th frame. $MAE$ indicates the accuracy of the estimates.

\subsection{Results and discussion}
To demonstrate the effectiveness of our proposed method, we compare our results to three methods: one that based on traditional machine learning method~\cite{he2017double}, and test another two deep learning based~\cite{zhang2016}~\cite{li2018csrnet} on subway station pedestrian dataset. Results of the extensive experiments are reported in Table 1. It can be observed that the proposed dynamic region learning algorithm obtains the lowest average MAE on the test frames. It is notable that during the training and testing process of ~\cite{he2017double}, background mask are used to effective features representation on ROI(region of interest). However, we don't use such background mask and process the whole frame in our method. Therefore, it can demonstrate that dynamic division can reserve more complete information than a straight line~\cite{he2017double} and density-based deep representation is more effective than handcrafted features in the distant region. At the same time, we can also observe that our method does not achieve the lowest $MAE$ in Jing'an temple and South railway station scene. We analyze these scenes and found that pedestrians are more likely to be occluded by other pedestrians and have larger occlusion area.

The counting and distribution estimation results are shown in Figure 5. The first column shows curve for test frames in each scene. The second column shows one sample image and the ground truth pedestrian count. The third row column shows the corresponding estimation result. It can be observed that our estimation curve is close to ground truth in most cases which means our method meet the requirement of public management.

\section{Conclusion}
In this paper, we propose an dynamic region learning algorithm for pedestrian counting in subway surveillance videos. The novel dynamic region division can meet the challenge of perspective distortion and avoid to cut the head into two parts. In the nearby region, we retrain YOLOV3 detector to substitute for using inaccurate gaussian kernels to model large scale heads. In the distant region, inception modules are used to automatically choose the dilation rate and achieve a better performance. The final fusion results can obtain more accurate counting result than methods before.

\bibliographystyle{IEEEbib}
\bibliography{icme2019template}

\end{document}